\documentclass[runningheads]{llncs}
\usepackage{graphicx}
\usepackage{amsmath,amssymb} 
\usepackage{color}
\usepackage{enumitem}
\usepackage[width=122mm,left=12mm,paperwidth=146mm,height=193mm,top=12mm,paperheight=217mm]{geometry}
\begin{document}
\pagestyle{headings}
\mainmatter
\def\ECCV16SubNumber{009}  

\title{Action-Affect Classification and Morphing\\ using Multi-Task Representation Learning} 

\titlerunning{Action-Affect Classification and Morphing}

\authorrunning{Timothy J. Shields}

\author{Timothy J. Shields\thanks{Both authors equally contributed to this work}, Mohamed R. Amer$^\star$, Max Ehrlich, Amir Tamrakar}
\institute{SRI International} 

\maketitle
\begin{abstract}
Most recent work focused on affect from facial expressions, and not as much on body. This work focuses on body affect analysis. Affect does not occur in isolation. Humans usually couple affect with an action in natural interactions; for example, a person could be talking and smiling. Recognizing body affect in sequences requires efficient algorithms to capture both the micro movements that differentiate between happy and sad and the macro variations between different actions. We depart from traditional approaches for time-series data analytics by proposing a multi-task learning model that learns a shared representation that is well-suited for action-affect classification as well as generation. For this paper we choose Conditional Restricted Boltzmann Machines to be our building block. We propose a new model that enhances the CRBM model with a factored multi-task component to become Multi-Task Conditional Restricted Boltzmann Machines (MTCRBMs). We evaluate our approach on two publicly available datasets, the Body Affect dataset and the Tower Game dataset, and show superior classification performance improvement over the state-of-the-art, as well as the generative abilities of our model.
\end{abstract}
\section{Introduction} \label{sec:Intro}
\begin{figure*}[t]
	\centering
	\includegraphics[width=\textwidth]{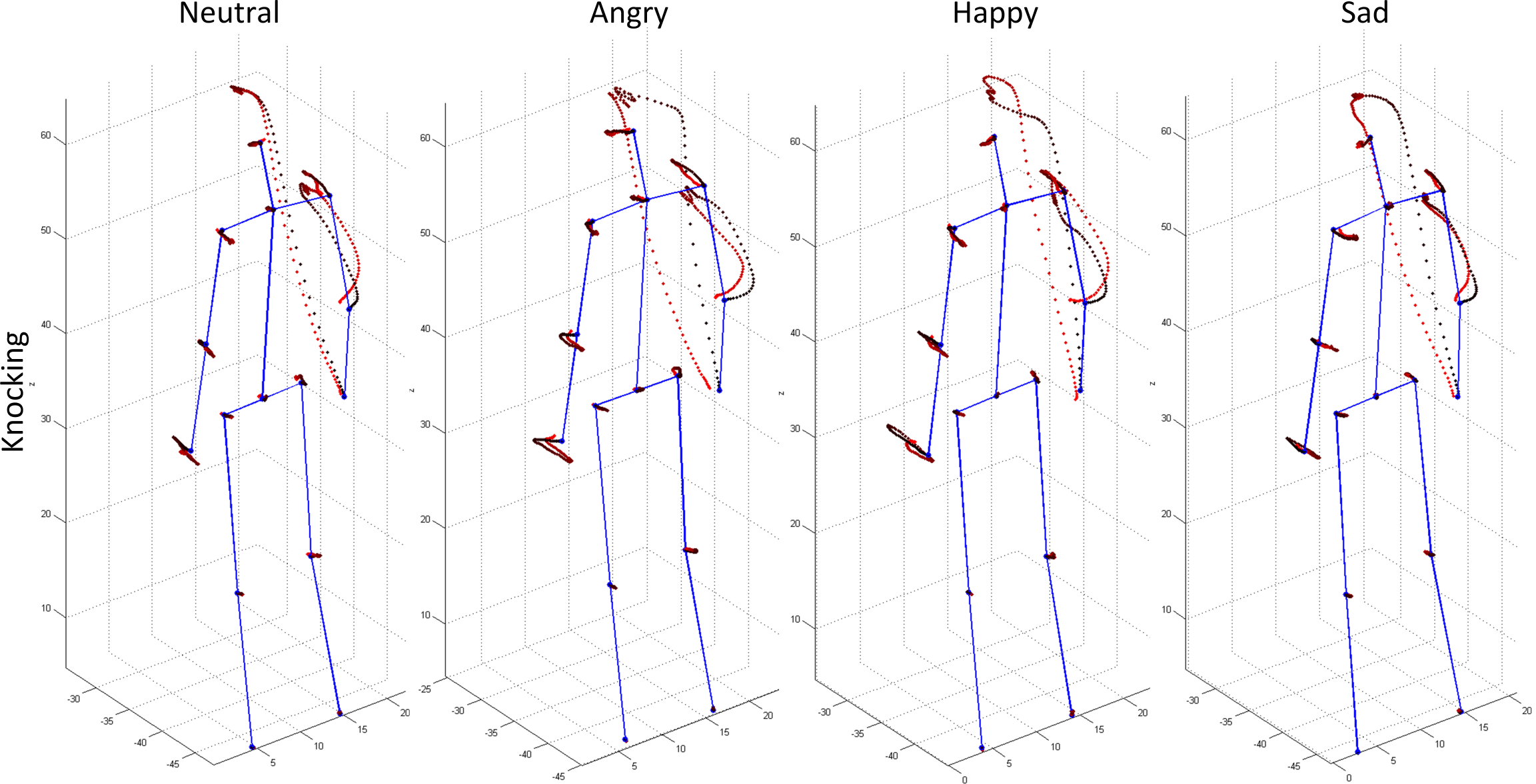}
	\caption{Examples from the Body Affect dataset \cite{Ma_BRM2006} of a person Knocking with various affects: Neutral, Angry, Happy, and Sad. The trajectory color corresponds to time, where black is the beginning of the sequence, reddish-black is the middle, and red is the end of the sequence. (This figure is best visualized in color.)}
	\label{fig:Intro}
\end{figure*}

There has been so much activity in the field of affective computing that it already contributed to the creation of new research directions in affect analysis \cite{Picard_MIT1995}. There are multiple research directions for analyzing human affect, including face data \cite{Calvo_OUP2014}, audiovisual data \cite{Zeng_PAMI2009}, and body data \cite{Kleinsmith_TAC2013}. One of the main challenges of affect analysis is that it does not occur in isolation. Humans usually couple affect with an action in natural interactions; for example, a person could be talking and smiling, or knocking on a door angrily as shown in Fig.~\ref{fig:Intro}. To be able to recognize body action-affect pairs, efficient temporal algorithms are needed to capture the micro movements that differentiate between happy and sad as well as capture the macro variations between the different actions. The focus of our work is on single-view, multi-task action-affect recognition from skeleton data captured by motion capture or Kinect sensors. Our work leverages the knowledge and work done by the graphics and animation community 
\cite{Amaya_GI1996,Rose_CGA1998,Ma_BRM2006} and uses machine learning to enhance it and make it accessible for a wide variety of applications. We use the Body Affect dataset produced by \cite{Ma_BRM2006} and the Tower Game \cite{Salter_ACII2015} dataset as the test cases for our novel multi-task approach.
Time series analysis is a difficult problem that requires efficient modeling, because of the large amounts of data it introduces. There are multiple approaches that designed features to reduce the data dimensionality, using mid-level features, and then use a simpler model to do classification \cite{Zhang_arxiv2016,Chaquet_CVIU2013}. We depart from these methods by proposing a model that learns a shared representation using multi-task learning. For this paper we choose Conditional Restricted Boltzmann Machines, which are non-linear generative models for modeling time series data, as our building block. They use an undirected model with binary latent variables connected to a number of visible variables. A CRBM-based generative model enables modeling short-term phenomenon. We propose a new hybrid model that enhances the CRBM model with multi-task, discriminative, components based on the work of \cite{Larochelle_ICML2008}. This work leads to a superior classification performance, while also allowing us to model temporal dynamics efficiently. We evaluate our approach on the Body Affect \cite{Ma_BRM2006} and Tower Game \cite{Salter_ACII2015} datasets and show how our results are superior to the state-of-the-art.\\

\noindent{\it Our contributions:} 
\begin{itemize}
	\item Multi-task learning model for unimodal and multimodal time-series data.
	\item Method for applying affect to a neutral skeleton (Sequence Morphing).
	\item Evaluations on two multi-task public datasets \cite{Ma_BRM2006,Salter_ACII2015}.
\end{itemize}

\noindent{\it Paper organization:} In sec.~\ref{sec:LitReview} we discuss prior work.  In sec.~\ref{sec:Model} we give a brief background of similar models that motivate our approach, followed by a description of our model. In sec.~\ref{sec:Inference} we describe the inference algorithm. In sec.~\ref{sec:Learning} we specify our learning algorithm. In sec.~\ref{sec:Experiments} we show quantitative results of our approach, followed by the conclusion in sec.~\ref{sec:Conclusion}.
\section{Prior work} \label{sec:LitReview}
In this section we first review literature on activity recognition in RGB-D and Motion Capture Sequences; second we review Multi-Task Learning approaches; finally we review temporal, energy-based, representation learning.\\

\noindent{\bf Body Affect Analysis:} Initial work on activity recognition in RGB-D sequences has been popular in recent years with the availability of cheap depth sensors. Since initial work \cite{Li_CVPRW2010}, there have been an increasing number of approaches addressing the problem of activity recognition using skeletal data \cite{Zhang_arxiv2016}. Prior to activity recognition in RGB-D sequences, datasets were captured using motion capture sensors. During that time, research focused on graphics applications such as generating animation and transitions between animations using signal processing techniques rather than machine learning or computer vision. Their main goal was to generate natural looking skeletons for animation. Some methods used knowledge of signal processing to transform a neutral skeleton pose to reflect a certain emotion \cite{Amaya_GI1996}. These methods were very constrained to the type of motion and were engineered to reproduce the same motions. Other work used a language based modeling of affect \cite{Rose_CGA1998} where they modeled actions (verbs) and affect (adverbs) using a graph. They were able to produce results using a combination of low level functions to interpolate between example motions. More recent work \cite{Bernhardt_ACII2007} modeling non-stylized motion for affect communication used segmentation techniques which divided complex motions into a set of motion primitives that they used as dynamic features. Unlike our approach, their mid-level features were hand engineered rather than learned, which is very limited, does not scale and is prone to feature design flaws. More recent work such as \cite{Ma_BRM2006} collected natural body affect datasets where they have varied identity, gender, emotion, and actions of the actors but not used it for classification.\\

\noindent{\bf Multi-Task Learning:} Multi-task learning is a natural approach for problems that require simultaneous solutions of several related problems \cite{Caruana_ML1997}. Multi-task learning approaches can be grouped into two main sets. The first set focuses on regularizing the parameter space. The main assumption is that there is an optimal shared parameter space for all tasks. These approaches regularize the parameter space by using a specific loss \cite{Evgeniou_KDD2004}, methods that manually define relationships \cite{Evgeniou_JMLR2005}, or more automatic ways that estimate the latent structure of relationships between tasks \cite{Ciliberto_CVPR2015,Maurer_arxiv2015,Maurer_ICML2013,Kumar_ICML2012,Zhou_NIPS2011}. The second set focuses on correlating relevant features jointly \cite{Argyriou_ML2008,Kang_ICML2011,Romera_ICML2013,Yang_ICLR2015}. Other work focused on the schedule of which tasks should be learned \cite{Pentina_CVPR2015}. Multi-task learning achieved good results on vision problems such as: person re-identification \cite{Su_ICCV2015}, multiple attribute recognition \cite{Chen_CVPR2014}, and tracking \cite{Zhang_PAMI2012}. Recently, Deep Multi-Task Learning (DMTL) emerged with the rise of deep learning. Deep Neural Networks (DNNs) were used to address multi-task learning and were applied successfully to facial landmark detection \cite{Zhang_ECCV2014}, scene classification \cite{Lapin_CVPR2014}, object localization and segmentation \cite{Wang_NIPS2014} and attribute prediction \cite{Abdulnabi_arxiv2016}. Other work used multi-task autoencoders \cite{Zhuang_ICML2015} for object recognition in a generalized domain \cite{Ghifary_ICCV2015}, where the tasks were the different domains. Other work used multi-task RNNs for interaction prediction in still images \cite{Chu_ICCV2015}. Most of the Deep Multi-task Learning approaches only focused on using DNN-based models applied to still images. Our approach is the first DMTL for temporal and multimodal sequence analysis.\\

\noindent{\bf Representation Learning:} Deep learning has been successfully applied to many problems \cite{Bengio_FTML2009}. Restricted Boltzmann Machines (RBMs) form the building blocks in energy-based deep networks \cite{Hinton_NC2006,Salakhutdinov_Science2006}. In \cite{Hinton_NC2006,Salakhutdinov_Science2006}, the networks are trained using the Contrastive Divergence (CD) algorithm \cite{Hinton_NC2002}, which demonstrated the ability of deep networks to capture the distributions over the features efficiently and to learn complex representations. RBMs can be stacked together to form deeper networks known as Deep Boltzmann Machines (DBMs), which capture more complex representations. Recently, temporal models based on deep networks have been proposed, capable of modeling a rich set of time series analysis problems. These include Conditional RBMs (CRBMs) \cite{Taylor_JMLR2011} and Temporal RBMs (TRBMs) \cite{Sutskever_AISTATS2007,Sutskever_NIPS2008,Hausler_CoRR2012}. CRBMs have been successfully used in both visual and audio domains. They have been used for modeling human motion \cite{Taylor_JMLR2011}, tracking 3D human pose \cite{Taylor_CVPR2010}, and phone recognition \cite{Mohamed_ICML2009}. TRBMs have been applied for transferring 2D and 3D point clouds \cite{Zeiler_NIPS2011}, and polyphonic music generation \cite{Lewandowski_ICML2009}.
\section{Model}\label{sec:Model}
Rather than immediately defining our Multi-Task CRBM (MT-CRBM) model, we discuss a sequence of models, gradually increasing in complexity, such that the different components of our final model can be understood in isolation. We start with the basic RBM model (sec.~\ref{sec:RBM}), then we extend the RBM to the CRBM model (sec.~\ref{sec:CRBM}), then we further extend the CRBM to a new discriminative (D-CRBM) model (sec.~\ref{sec:DCRBM}), then we extend the D-CRBM to our main multi-task model (MT-CRBM) (sec.~\ref{sec:MTCRBM}), and finally we define a multi-task multimodal model (MTM-CRBM) (sec.~\ref{sec:MTMCRBM}).
\begin{figure*}[t]
	\centering
	\begin{minipage}{0.32\textwidth}
		\vspace{28pt}
		\centering
		\includegraphics[width=\textwidth]{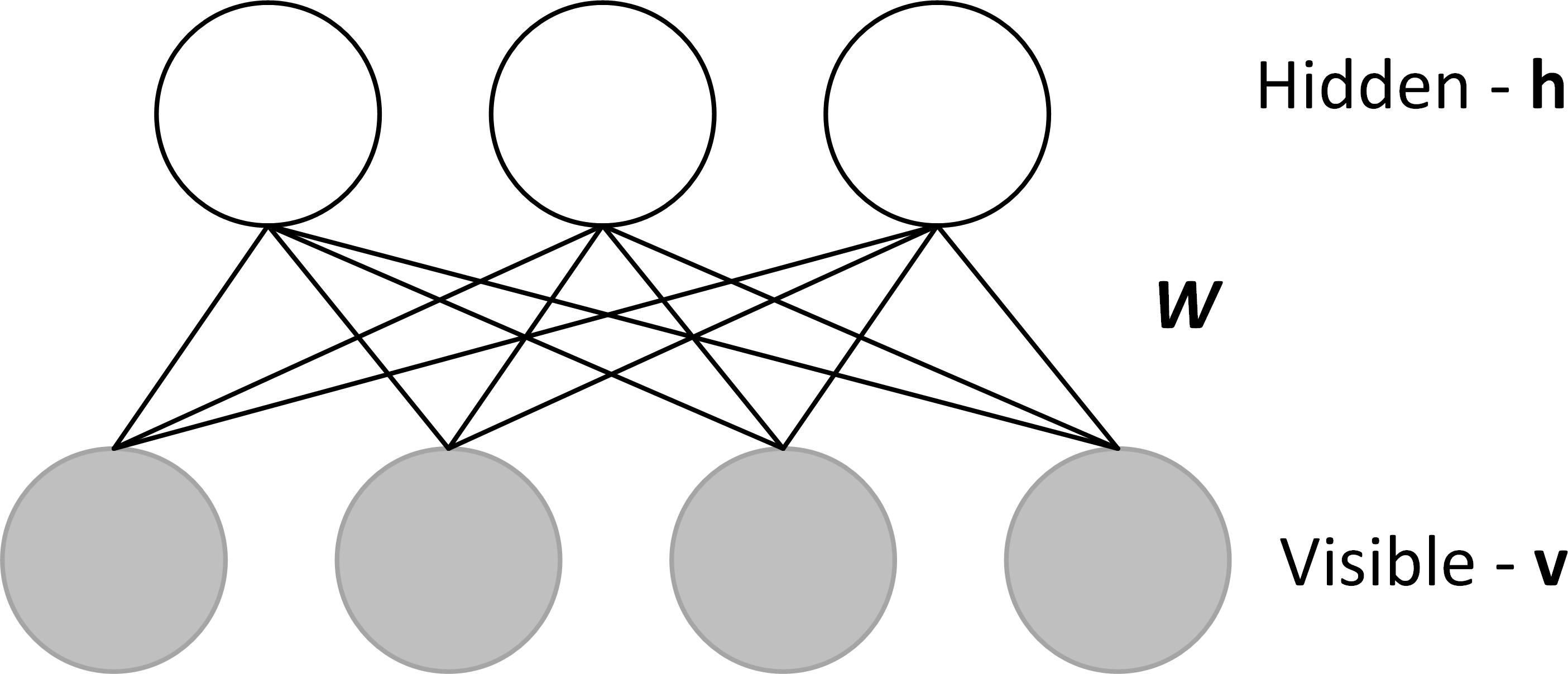}
		\scriptsize{(a) RBM}		
	\end{minipage}
	\hfill%
	\begin{minipage}{0.32\textwidth}
		\vspace{30pt}
		\centering
		\includegraphics[width=\textwidth]{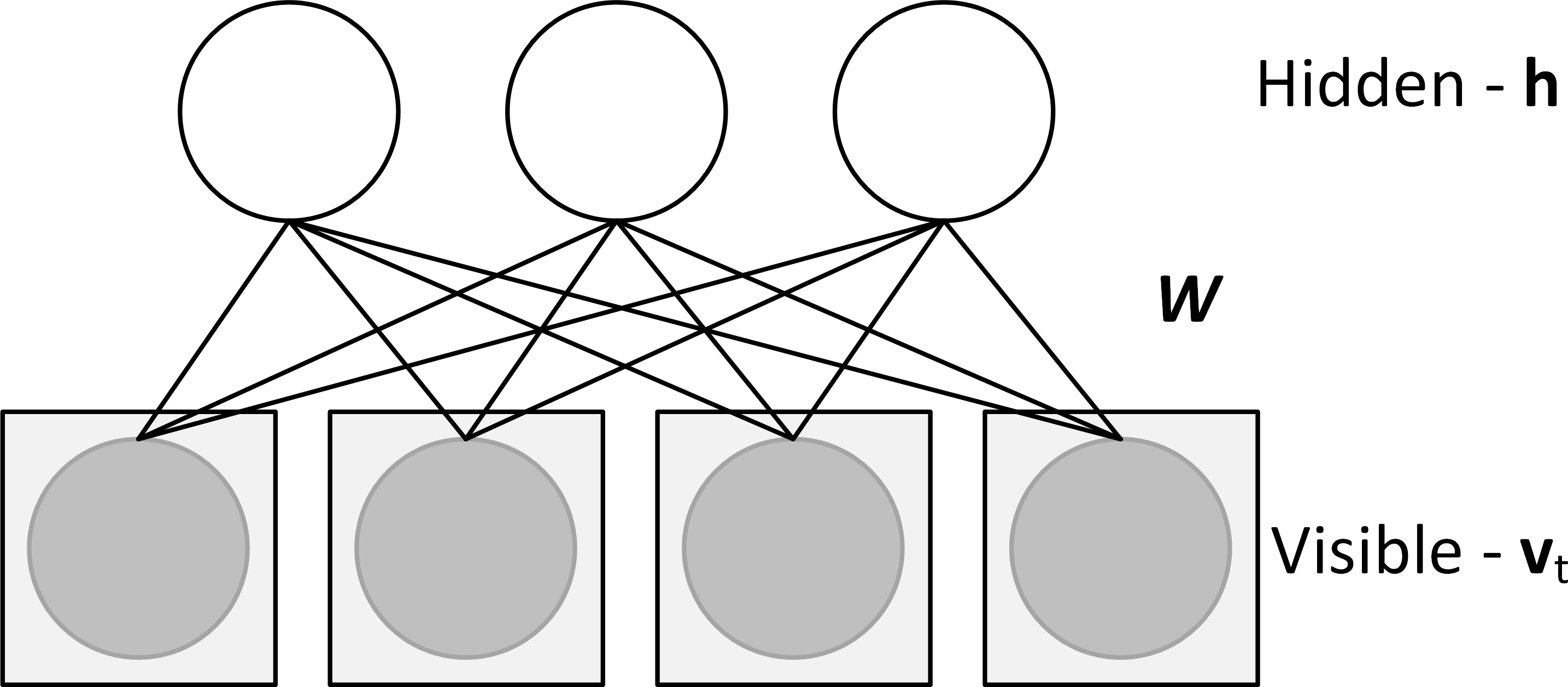}
		\scriptsize{(b) CRBM}		
	\end{minipage}
	\hfill%
	\begin{minipage}{0.32\textwidth}
		\centering
		\includegraphics[width=\textwidth]{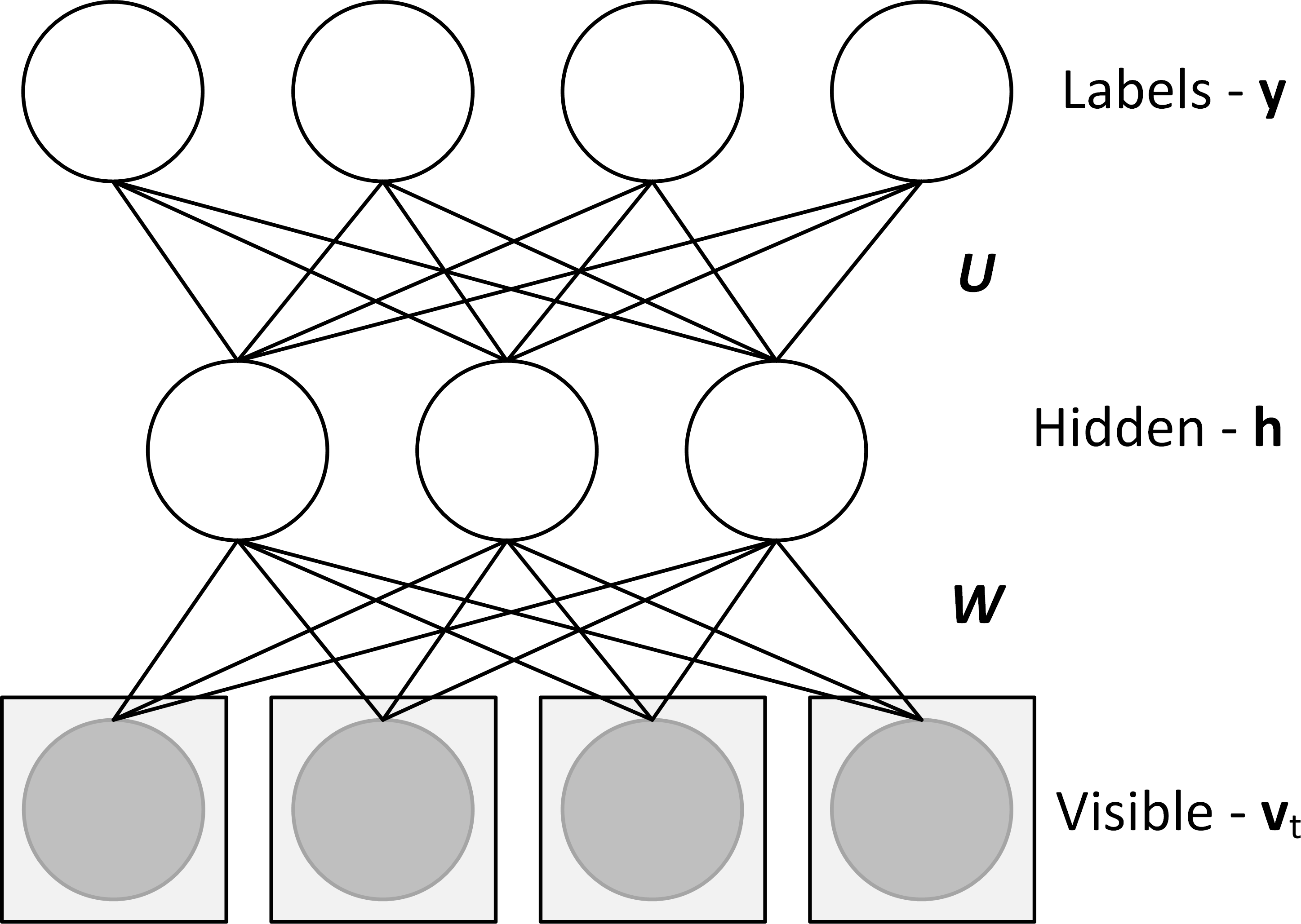}
		\scriptsize{(c) D-CRBM}		
	\end{minipage}
	\caption{The deep learning models described in sections~\ref{sec:RBM}, \ref{sec:CRBM}, and \ref{sec:DCRBM}: (a)~RBM (b)~CRBM (c)~D-CRBM.}
	\label{fig:RCD}
\end{figure*}
\subsection{Restricted Boltzmann Machines}\label{sec:RBM}
RBMs \cite{Salakhutdinov_Science2006}, shown in Figure~\ref{fig:RCD}(a), define a probability distribution $p_{\text{R}}$ as a Gibbs distribution (\ref{eqn:RBM}), where ${\bf v}$ is a vector of visible nodes, ${\bf h}$ is a vector of hidden nodes, $E_{\text{R}}$ is the energy function, and $Z$ is the partition function. The parameters ${\boldsymbol \theta}$ to be learned are ${\bf a}$ and ${\bf b}$, the biases for ${\bf v}$ and ${\bf h}$ respectively, and the weights ${\it W}$. The RBM is fully connected between layers, with no lateral connections.  This architecture implies that {\bf v} and {\bf h} are factorial given one of the two vectors.  This allows for the exact computation of $p_{\text{R}}({\bf v}|{\bf h})$ and $p_{\text{R}}({\bf h}|{\bf v})$.
\begin{equation}
\centering
\begin{array}{c}
p_{\text{R}}({\bf h},{\bf v})=\frac{\exp[-E_{\text{R}}({\bf h},{\bf v})]}{Z({\boldsymbol \theta})},\\
\\
Z({\boldsymbol \theta})=\sum_{{\bf h},{\bf v}}\exp[-E_{\text{R}}({\bf h},{\bf v})],
\end{array}
\quad {\boldsymbol \theta}=
\Bigg[
\begin{matrix}
\{{\bf a},{\bf b}\}&\text{-bias},\\
\{{\it W}\}&\text{-fully connected}.\\
\end{matrix}
\Bigg]
\label{eqn:RBM}
\end{equation}
In case of binary valued data $v_i$ is defined as  a logistic function. In case of real valued data, $v_i$ is defined as a multivariate Gaussian distribution with a unit covariance. A binary valued hidden layer $h_j$ is defined as a logistic function such that the hidden layer becomes sparse \cite{Taylor_JMLR2011,Salakhutdinov_AISTATS2009}. The probability distributions over $v$ and $h$ are defined as in (\ref{eqn:PRBM}).
\begin{equation}
\begin{array}{rcll}
p_{\text{R}}(v_{i} = 1 |{\bf h})&=&\sigma(a_{i}+\sum_{j} h_{j} w_{ij}),\quad& \text{Binary,}\\
\\
p_{\text{R}}(v_{i}|{\bf h})&=&\mathcal{N}(a_{i}+\sum_{j} h_{j}w_{ij},1),\quad& \text{Real,}\\
\\
p_{\text{R}}(h_{j} = 1 |{\bf v})&=&\sigma(b_{j}+\sum_{i} v_{i} w_{ij}),\quad& \text{Binary.}
\end{array}
\label{eqn:PRBM}
\end{equation} 
The energy function $E_{\text{R}}$ for the real valued ${\bf v}$ is defined as in (\ref{eqn:ERBM}).
\begin{equation}
E_{\text{R}}({\bf h},{\bf v})=\sum_{i} \frac{(a_{i}-v_{i})^2}{2} - \sum_{j} b_{j} h_{j}- \sum_{i,j} v_{i}w_{ij} h_{j}
\label{eqn:ERBM}
\end{equation}
\subsection{Conditional Restricted Boltzmann Machines}\label{sec:CRBM}
CRBMs \cite{Taylor_JMLR2011} are a natural extension of RBMs for modeling short term temporal dependencies. A CRBM, shown in Figure~\ref{fig:RCD}(b), is an RBM which takes into account history from the previous $N$ time instances, $t-N,\hdots,t-1$, when considering time $t$. This is done by treating the previous time instances as additional inputs. Doing so does not complicate inference. Some approximations have been made to facilitate efficient training and inference, more details are available in \cite{Taylor_JMLR2011}. A CRBM defines a probability distribution $p_{\text{C}}$ as a Gibbs distribution (\ref{eqn:CRBM}). 
\begin{equation}
	\centering
	\begin{array}{c}
		p_{\text{C}}({\bf h}_{t},{\bf v}_{t}|{\bf v}_{<t})=\frac{\exp[-E_{\text{C}}({\bf v}_{t},{\bf h}_{t}|{\bf v}_{<t})]}{Z({\boldsymbol \theta})},\\
		\\
		Z({\boldsymbol \theta})=\sum_{{\bf h},{\bf v}}\exp[-E_{\text{C}}({\bf h}_{t},{\bf v}_{t}|{\bf v}_{<t})],
	\end{array}
	\quad	{\boldsymbol \theta}=
	\Bigg[
	\begin{matrix}
	\{{\bf a},{\bf b}\}&\text{-bias},\\
	\{{\it A},{\it B}\}&\text{-auto regressive},\\
	\{{\it W}\}&\text{-fully connected}.\\
	\end{matrix}
	\Bigg]
	\label{eqn:CRBM}
\end{equation}
The visible vectors from the previous $N$ time instances, denoted as ${\bf v}_{<t}$, influence the current visible and hidden vectors. The probability distributions are defined in (\ref{eqn:PCRBM}).
\begin{equation}
	\begin{array}{rcl}
		p_{\text{C}}(v_{i}|{\bf h},{\bf v}_{<t})&=&\mathcal{N}(c_i+ \sum_{j} h_{j}w_{ij},1),\\
		\\
		p_{\text{C}}(h_{j} = 1 |{\bf v},{\bf v}_{<t})&=&\sigma(d_j + \sum_{i} v_{i} w_{ij}),\\
		\\
		c_{i}= a_{i} + \sum_{p}A_{pi} v_{p,<t}&,\quad& d_{j}=b_{j} + \sum_{p}B_{pj} v_{p,<t}.
	\end{array}
	\label{eqn:PCRBM}
\end{equation}
The new energy function $E_{\text{C}}({\bf h}_{t},{\bf v}_{t}|{\bf v}_{<t})$ in (\ref{eqn:ECRBM}) is defined in a manner similar to that of the RBM (\ref{eqn:ERBM}).
\begin{equation}
	\begin{array}{c}
		E_{\text{C}}({\bf h}_{t},{\bf v}_{t}|{\bf v}_{<t})=\sum_{i} (c_{i}-v_{i,t})^2/2 - \sum_{j} d_{j} h_{j,t}- \sum_{i,j} v_{i,t} w_{ij} h_{j,t},
	\end{array}
	\label{eqn:ECRBM}
\end{equation}
Note that $A$ and $B$ are matrices defining dynamic biases for ${\bf v}_t$ and ${\bf h}_t$, consisting of concatenated vectors of previous time instances of ${\bf a}$ and ${\bf b}$.
\subsection{Discriminative Conditional Restricted Boltzmann Machines}\label{sec:DCRBM}
We extend the CRBMs to the D-CRBMs shown in Figure~\ref{fig:RCD}(c). D-CRBMs are based on the D-RBM model presented in in \cite{Larochelle_ICML2008}, generalized to account for temporal phenomenon using CRBMs. D-CRBMs define the probability distribution $p_{\text{DC}}$ as a Gibbs distribution (\ref{eqn:DCRBM}). 
\begin{equation}
\begin{array}{rcl}
p_{\text{DC}}({\bf y}_{t},{\bf h}_{t},{\bf v}_{t}|{\bf v}_{<t})=\frac{\exp[-E_{\text{DC}}( {\bf y}_{t},{\bf h}_{t},{\bf v}_{t}|{\bf v}_{<t})]}{Z({\boldsymbol \theta})},\\
\\
Z({\boldsymbol \theta})=\sum_{{\bf y},{\bf h},{\bf v}}\exp[-E_{\text{DC}}({\bf y}_{t},{\bf h}_{t},{\bf v}_{t}|{\bf v}_{<t})],
\end{array}
\quad {\boldsymbol \theta}=
\Bigg[
\begin{matrix}
\{{\bf a},{\bf b},{\bf s}\}&\text{-bias},\\
\{{\it A},{\it B}\}&\text{-auto regressive},\\
\{{\it W},{\it U}\}&\text{-fully connected}.
\end{matrix}
\Bigg]
\label{eqn:DCRBM}
\end{equation}
The probability distribution over the visible layer will follow the same distributions as in (\ref{eqn:PCRBM}). The hidden layer ${\bf h}$ is defined as a function of the labels $y$ and the visible nodes ${\bf v}$. A new probability distribution for the classifier is defined to relate the label $y$ to the hidden nodes ${\bf h}$ (\ref{eqn:PDCRBM}).
\begin{equation}
\begin{array}{rcl}
p_{\text{DC}}(v_{i,t}|{\bf h}_{t},{\bf v}_{<t})&=&\mathcal{N}(c_i + \sum_{j} h_{j}w_{ij},1),\\
\\
p_{\text{DC}}(h_{j,t} = 1 |y_{t},{\bf v}_{t},{\bf v}_{<t})&=& \sigma(d_j+ \sum_k y_{k,t} u_{jk} + \sum_{i} v_{i,t} w_{ij}),\\
\\
p_{\text{DC}}(y_{k,t}|{\bf h})&=&\frac{\exp[s_k+\sum_j u_{jk}h_j]}{\sum_{k^*}\exp[s_{k^*}+\sum_j u_{jk^*}h_j]}.
\end{array}
\label{eqn:PDCRBM}
\end{equation}
The new energy function $E_{\text{DC}}$ is defined as in (\ref{eqn:EDCRBM}).
\begin{equation}
E_{\text{DC}}({\bf y}_{t},{\bf h}_{t},{\bf v}_{t}|{\bf v}_{<t})= \underbrace{E_{\text{C}}({\bf h}_{t},{\bf v}_{t}|{\bf v}_{<t})}_{\text{Generative}} - \underbrace{\sum_{k} s_{k} y_{k,t}-\sum_{j,k} h_{j,t} u_{jk} y_{k,t}}_{\text{Discriminative}}
\label{eqn:EDCRBM}
\end{equation}
\begin{figure*}[t]
	\centering
	\begin{minipage}{0.49\textwidth}
		\centering
		\includegraphics[width=0.96\textwidth]{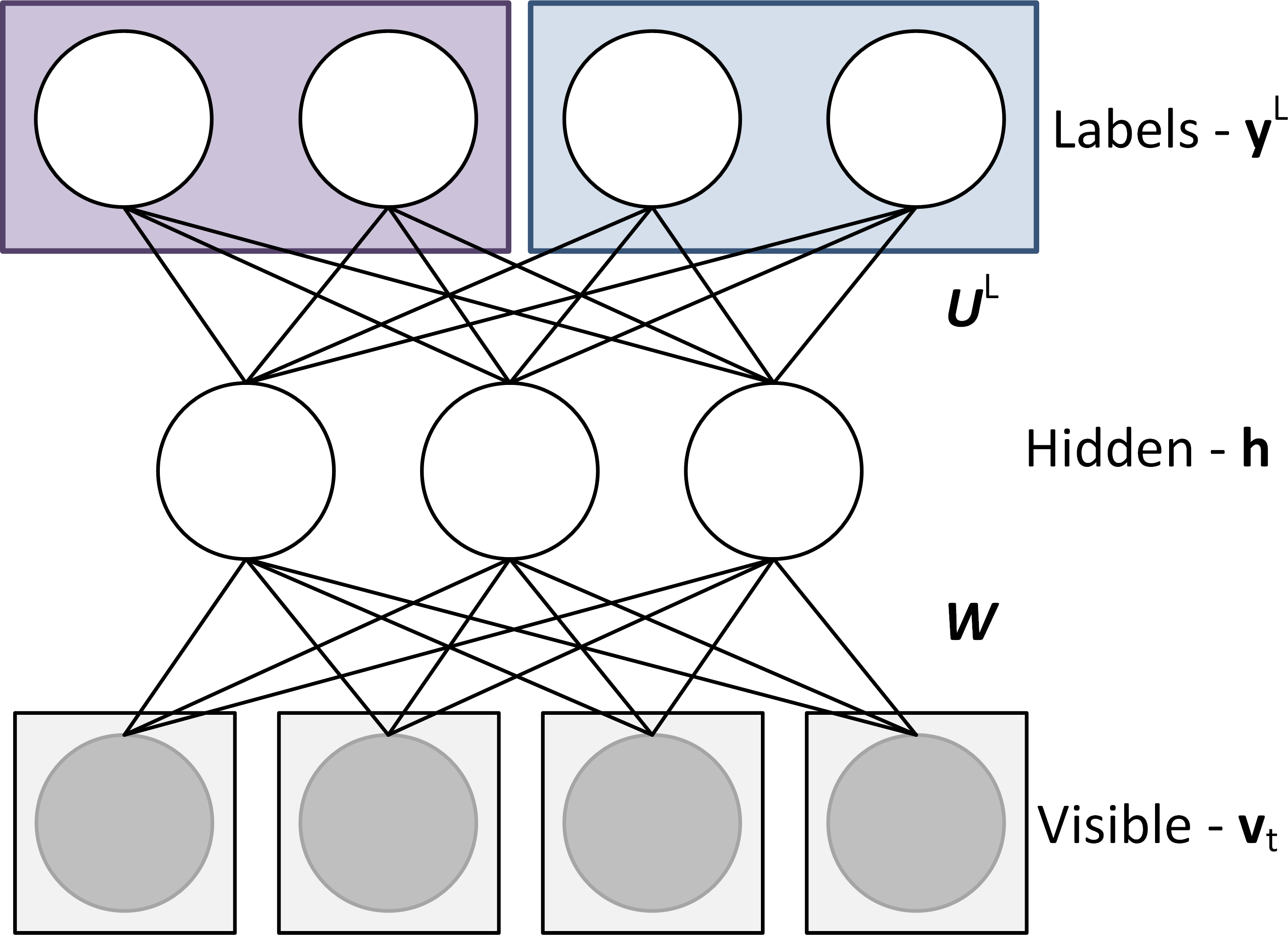}\\
		\scriptsize{(a) MT-CRBMs}
		\vspace{46pt}
	\end{minipage}	
	\hfill%
	\begin{minipage}{0.49\textwidth}
		\centering		
		\includegraphics[width=0.98\textwidth]{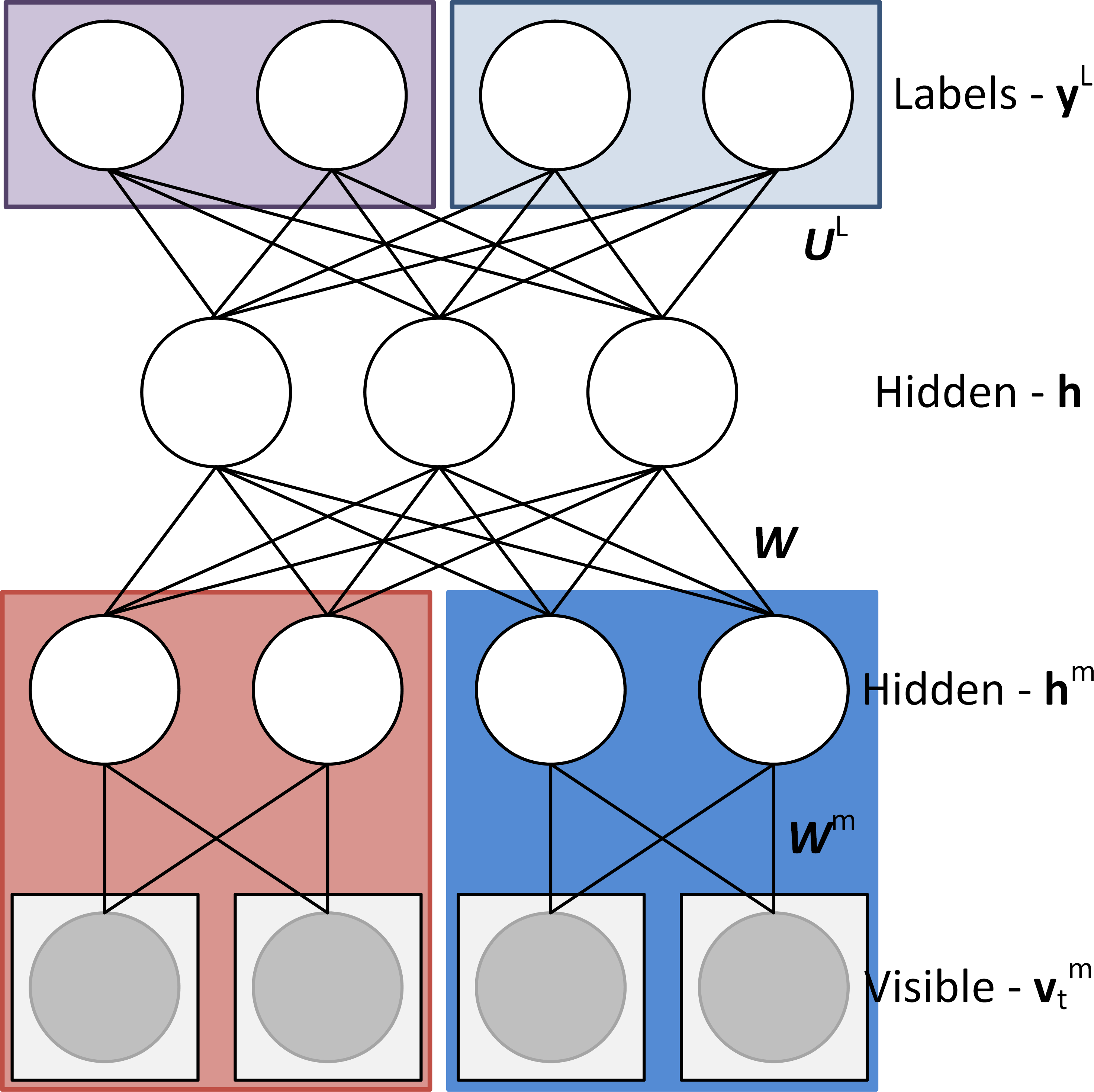}\\
		\scriptsize{(b) MTM-CRBMs}
	\end{minipage}	
	\caption{Models described in sections~\ref{sec:MTCRBM} and ~\ref{sec:MTMCRBM}. (a) MT-CRBMs (b) MTM-CRBMs. The MT-CRBMs learn a shared representation layer for all tasks. In addition to the shared layer, the MTM-CRBMs learn an extra representation layer for each of the modalities, which learn modality-specific representations.}
	\label{fig:MT}
\end{figure*}
\subsection{Multi-Task Conditional Restricted Boltzmann Machines}\label{sec:MTCRBM}
In the same way the CRBMs can be extended to the DC-RBMs by adding a discriminative term to the model, we can extend the CRBMs to be multi-task MT-CRBMs Figure~\ref{fig:MT}(a). MTCRBMs define the probability distribution $p_{\text{MT}}$ as a Gibbs distribution (\ref{eqn:MTCRBM}). The MT-CRBMs learn a shared representation layer for all tasks.
\begin{equation}
\begin{array}{l}
p_{\text{MT}}({\bf y}^{L}_{t},{\bf h}_{t},{\bf v}_{t}|{\bf v}_{<t})=\frac{\exp[-E_{\text{DC}}( {\bf y}^{L}_{t},{\bf h}_{t},{\bf v}_{t}|{\bf v}_{<t})]}{Z({\boldsymbol \theta})},\\
\\
Z({\boldsymbol \theta})=\sum_{{\bf y},{\bf h},{\bf v}}\exp[-E_{\text{MT}}({\bf y}^{L}_{t},{\bf h}_{t},{\bf v}_{t}|{\bf v}_{<t})],
\end{array}
\quad{\boldsymbol \theta}=
\Bigg[
\begin{matrix}
\{{\bf a},{\bf b},{\bf s}^{L}\}&\text{-bias},\\
\{{\it A},{\it B}\}&\text{-auto regressive},\\
\{{\it W},{\it U}^{L}\}&\text{-fully connected}.
\end{matrix}
\Bigg]
\label{eqn:MTCRBM}
\end{equation}
The probability distribution over the visible layer will follow the same distributions as in (\ref{eqn:PDCRBM}). The hidden layer ${\bf h}$ is defined as a function of the multi-task labels $y^L$ and the visible nodes ${\bf v}$. A new probability distribution for the multi-task classifier is defined to relate the multi-task labels $y^L$ to the hidden nodes ${\bf h}$ as shown in (\ref{eqn:PMTCRBM}).
\begin{equation}
\begin{array}{rcl}
p_{\text{MT}}(v_{i,t}|{\bf h}_{t},{\bf v}_{<t})&=&\mathcal{N}(c_i + \sum_{j} h_{j}w_{ij},1),\\
\\
p_{\text{MT}}(h_{j,t} = 1 |y^{L}_{t},{\bf v}_{t},{\bf v}_{<t})&=&\sigma(d_j+ \sum_{l,k} y^{l}_{k,t} u^{l}_{jk} + \sum_{i} v_{i,t} w_{ij}),\\
\\
p_{\text{MT}}(y^{l}_{k,t}|{\bf h})&=&\frac{\exp[s^{l}_k+\sum_j u^{l}_{jk}h_j]}{\sum_{k^*}\exp[s^{l}_{k^*}+\sum_j u^{l}_{jk^*}h_j]}.
\end{array}
\label{eqn:PMTCRBM}
\end{equation}
The energy for the model shown in Figure~\ref{fig:MT}(a), $E_{\text{MT}}$, is defined as in (\ref{eqn:EMTCRBM}).
\begin{equation}
E_{\text{MT}}({\bf y}^{L}_{t},{\bf h}_{t},{\bf v}_{t}|{\bf v}_{<t})= \underbrace{E_{\text{C}}({\bf v}_{t},{\bf h}_{t}|{\bf v}_{<t})}_{\text{Generative}} - \underbrace{\sum_{k,l} s^{l}_{k} y^{l}_{k,t}-\sum_{j,k,l} h_{j,t} u_{jk} y^{l}_{k,t}}_{\text{Multi-Task}}
\label{eqn:EMTCRBM}
\end{equation}
\subsection{Multi-Task Multimodal Conditional Restricted Boltzmann Machines}\label{sec:MTMCRBM}
We can naturally extend MT-CRBMs to MTM-CRBMs. A MTM-CRBMs combines a collection of unimodal MT-CRBMs, one for each visible modality. The hidden representations produced by the unimodal MT-CRBMs are then treated as the visible vector of a single fusion MT-CRBMs. The result is a MTMCRBM model that relates multiple temporal modalities to multi-task classification labels. MTMCRBMs define the probability distribution $p_{\text{MTM}}$ as a Gibbs distribution (\ref{eqn:MTMCRBM}). The MTM-CRBMs learn an extra representation layer for each of the modalities, which learns a modality specific representation as well as the shared layer for all the tasks.
\begin{equation}
\begin{array}{c}
p_{\text{MTM}}({\bf y}^{L}_{t},{\bf h}_{t},{\bf h}^{1:M}_{t},{\bf v}^{1:M}_{t}|{\bf v}^{1:M}_{<t})=\exp[-E_{\text{MTM}}({\bf y}^{L}_{t},{\bf h}_{t},{\bf h}^{1:M}_{t},{\bf v}^{1:M}_{t}|{\bf v}^{1:M}_{<t})]/Z({\boldsymbol \theta}),\\
\\
Z({\boldsymbol \theta})=\sum_{{\bf y},{\bf v},{\bf h}}\exp[-E_{\text{MTM}}({\bf y}^{L}_{t},{\bf h}_{t},{\bf h}^{1:M}_{t},{\bf v}^{1:M}_{t}|{\bf v}^{1:M}_{<t}),\\
\\
{\boldsymbol \theta}=
\Bigg[
\begin{matrix}
\{{\bf a}^{1:M},{\bf b}^{1:M},{\bf e},{\bf s}^{L}\}	   &\text{-bias},\\
\{{\it A}^{1:M},{\it B}^{1:M},{\it C}^{1:M}\}	   &\text{-auto regressive},\\
\{{\it W}^{1:M},{\it U}^{1:M},{\it W},{\it U}^{L}\}  &\text{-fully connected}.
\end{matrix}
\Bigg]
\end{array}
\label{eqn:MTMCRBM}
\end{equation}
Similar to the MT-CRBMs(\ref{eqn:PMTCRBM}), the hidden layer ${\bf h}$ is defined as a function of the labels $y^{L}$ and the visible nodes ${\bf v}$. A new probability distribution for the classifier is defined to relate the label $y^{L}$ to the hidden nodes ${\bf h}$ is defined as in (\ref{eqn:PMTMCRBM}).
\begin{equation}
\begin{array}{l}
p_{\text{MTM}}(v^{m}_{i,t}|{\bf h}^{m}_{t},{\bf v}^{m}_{<t})=\mathcal{N}(c^{m}_{i} + \sum_{j} h^{m}_{j}w^{m}_{ij},1),\\
\\
p_{\text{MTM}}(h^{m}_{j,t} = 1 |y^{L}_{t},{\bf v}^{m}_{t},{\bf v}^{m}_{<t})= \sigma(d^{m}_{j} + \sum_{l,k} y^{l}_{k,t} u^{l}_{jk} + \sum_{i} v^{m}_{i,t} w^{m}_{ij}),\\
\\
p_{\text{MTM}}(y^{l}_{k,t}|{\bf h}_{t}^{m})=\frac{\exp[s^{l}_k+\sum_j u^{m,l}_{jk}h_{j,t}^{m}]}{\sum_{l^*} \exp[s^{l}_{k^*}+\sum_j u^{m,l}_{jk^*}h_{j,t}^{m}]},\\
\\
p_{\text{MTM}}(h_{n,t} = 1 |y^{L}_{t},{\bf h}^{1:M}_{t},{\bf h}^{1:M}_{<t})= \sigma(f_{n} + \sum_{l,k} y^{l}_{k,t} u^{l}_{nk} + \sum_{m,j} h^{m}_{j,t} w^{m}_{jn}),\\
\\
p_{\text{MTM}}(y^{l}_{k,t}|{\bf h})=\frac{\exp[s^{l}_k+\sum_j u^{l}_{nk}h_n]}{\sum_{k^*}\exp[s^{l}_{k^*}+\sum_{n} u^{l}_{nk^*}h_n]}.
\end{array}
\label{eqn:PMTMCRBM}
\end{equation}
where,
\begin{equation}
\begin{array}{rcccl}
c^{m}_{i} &=& a^{m}_{i} &+& \sum_{p}A^{m}_{p,i} v^{m}_{p,<t},\\
\\
d^{m}_{j} &=& b^{m}_{j} &+& \sum_{p}B^{m}_{p,j} v_{p,<t},\\
\\
f_{n} &=& e_{n} &+& \sum_{m,r}C_{r,n}^{m} h^{m}_{r,<t}.
\end{array}
\label{eqn:CDMTMCRBM}
\end{equation}
The new energy function $E_{\text{MTM}}$ is defined in (\ref{eqn:EMTMCRBM}) similar to that of the MT-CRBMs (\ref{eqn:MTCRBM}).
\begin{equation}
\begin{array}{c}
E_{\text{MTM}}({\bf y}^{L}_{t},{\bf h}_{t},{\bf h}^{1:M}_{t},{\bf v}^{1:M}_{t}|{\bf v}^{1:M}_{<t})= \underbrace{\sum_{m} E_{\text{MT}}({\bf y}^{L}_{t},{\bf h}^{m}_{t},{\bf v}^{m}_{t}|{\bf v}^{m}_{<t})}_{\text{Unimodal}} \\
\\
- \underbrace{\sum_{j} f_{n} h_{n,t}-\sum_{j,k,m}h^{m}_{j,t}w_{jn} h_{n,t}}_{\text{Fusion}} - \underbrace{\sum_{k,l} s^{l}_{k} y^{l}_{k,t}-\sum_{n,k,l} h_{n,t} u^{l}_{nk} y^{l}_{k,t}}_{\text{Multi-Task}}
\end{array}
\label{eqn:EMTMCRBM}
\end{equation}
%
%
%
%
\section{Inference}\label{sec:Inference}
\begin{figure*}
	\centering
	\begin{minipage}{0.49\textwidth}
		\centering
		\includegraphics[width=\textwidth]{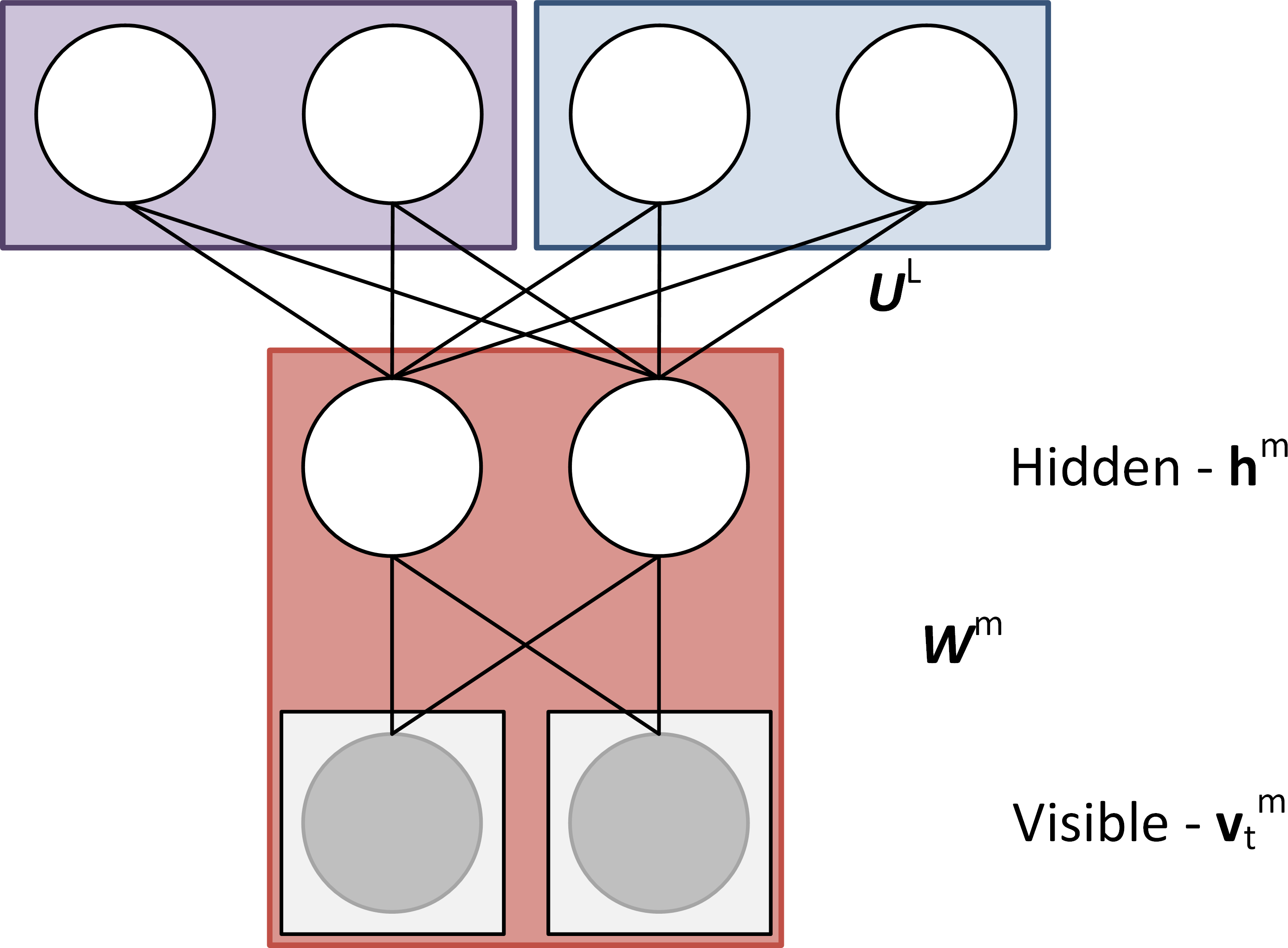}\\
		\scriptsize{(a) Unimodal MTCRBM}
	\end{minipage}
	\hfill%
	\begin{minipage}{0.49\textwidth}
		\centering
		\includegraphics[width=\textwidth]{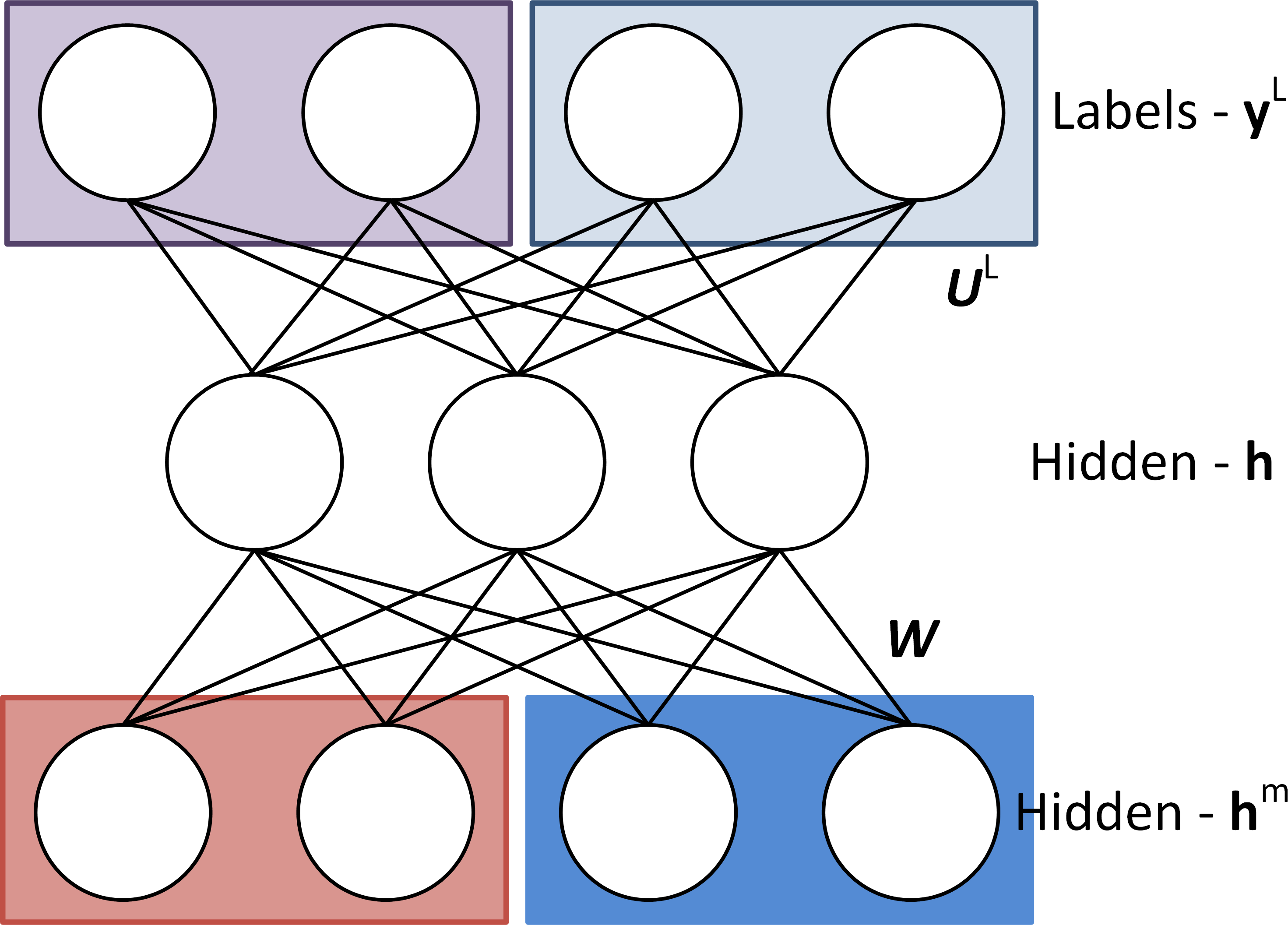}\\
		\scriptsize{(b) Fusion MTCRBM}
	\end{minipage}		
	\caption{This figure specifies the inference algorithm.  We first classify the unimodal data by activating the corresponding hidden layers ${\bf h}^{m}_{t}$ as shown in (a), followed by classifying the multimodal data by activating the fusion layer ${\bf h}_{t}$ as shown in (b).}
	\label{fig:InferenceAndLearningFigure}
\end{figure*}
We first discuss inference for the MTM-CRBM since it is the most general case. To perform classification at time $t$ in the MTM-CRBM given ${\bf v}_{<t}^{1:M}$ and ${\bf v}_{t}^{1:M}$ we use a bottom-up approach, computing the mean of each node given the activation coming from the nodes below it; that is, we compute the mean of ${\bf h}_t^m$ using ${\bf v}_{<t}^{m}$ and ${\bf v}_{t}^{m}$ for each modality, then we compute the mean of ${\bf h}_t$ using ${\bf h}_{<t}^{1:M}$, then we compute the mean of ${\bf y}^{L}_{t}$ for each task using ${\bf h}_t$, obtaining the classification probabilities for each task. Figure \ref{fig:InferenceAndLearningFigure} illustrates our inference approach. Inference in the MT-CRBM is the same as the MTM-CRBM, except there is only one modality, and inference in the D-CRBM is the same as the MT-CRBM, except there is only one task.
\section{Learning}\label{sec:Learning}
Learning our model is done using Contrastive Divergence (CD) \cite{Hinton_NC2002}, where $\langle\cdot\rangle _{data}$ is the expectation with respect to the data and $\langle\cdot\rangle _{recon}$ is the expectation with respect to the reconstruction. The learning is done using two steps: a bottom-up pass and a top-down pass using sampling equations from (\ref{eqn:PDCRBM}) for D-CRBM, (\ref{eqn:PMTCRBM}) for MT-CRBM, and (\ref{eqn:PMTMCRBM}) for MTM-CRBM. In the bottom-up pass the reconstruction is generated by first sampling the unimodal layers $p(h^{m}_{t,j}=1|{\bf v}_{t}^{m},{\bf v}^{m}_{<t},y_l)$ for all the hidden nodes in parallel. This is followed by sampling the fusion layer $p(h_{t,n}=1|y^{L}_{k,t},{\bf h}_{t}^{1:M},{\bf h}^{1:M}_{<t})$. In the top-down pass the unimodal layer is generated using the activated fusion layer $p(h^{m}_{t,j}=1|{\bf h}_{t},y^{L}_{k,t})$. This is followed by sampling the visible nodes  $p(v^{m}_{t,i}|{\bf h}_{t}^{m},{\bf v}^{m}_{<t})$ for all the visible nodes in parallel. The gradient updates are described in (\ref{eqn:CD}). Similarly learning of D-CRBM and MT-CRBM could be done.
\begin{equation}
	\centering
	\begin{array}{lclcl}
		\Delta a_{i}&\propto&\langle v^{m}_{i}\rangle _{data} &-& \langle v^{m}_{i}\rangle _{recon},\\
		\Delta b_{j}&\propto&\langle h^{m}_{j}\rangle _{data} &-& \langle h^{m}_{j}\rangle _{recon},\\
		\Delta e_{n}&\propto&\langle h_{n}\rangle _{data} &-& \langle h_{n}\rangle _{recon},\\
		\Delta s^{l}_{k}&\propto&\langle y^{l}_{k}\rangle _{data} &-& \langle y^{l}_{k}\rangle _{recon},\\ 
		\Delta A^{m}_{p,i,<t}&\propto& v^{m}_{k,<t}(\langle v^{m}_{i,t}\rangle _{data} &-& \langle v^{m}_{i,t}\rangle _{recon}),\\
		\Delta B^{m}_{p,j,<t}&\propto& v^{m}_{i,<t}(\langle h^{m}_{j,t}\rangle _{data} &-& \langle h^{m}_{j,t}\rangle _{recon}),\\
		\Delta C^{m}_{r,n,<t}&\propto& h^{m}_{j,<t}(\langle h_{n,t}\rangle _{data} &-& \langle h_{n,t}\rangle _{recon}),\\
		\Delta w^{m}_{i,j}&\propto&\langle v^{m}_{i}h^{m}_{j}\rangle _{data} &-& \langle v^{m}_{i}h^{m}_{j}\rangle _{recon},\\ 
		\Delta w_{j,k}&\propto&\langle h^{m}_{j}h_{n}\rangle _{data} &-& \langle h^{m}_{j}h_{n}\rangle _{recon},\\ 
		\Delta u^{l,m}_{jk}&\propto&\langle y^{l}_{k}h^{m}_{j}\rangle _{data} &-& \langle y^{l}_{k}h^{m}_{j}\rangle _{recon},\\ 
		\Delta u^{L}_{nk}&\propto&\langle y^{l}_{k}h_{n}\rangle _{data} &-& \langle y^{l}_{k}h_{n}\rangle _{recon}.
	\end{array}
	\label{eqn:CD}
\end{equation}
\section{Experiments}\label{sec:Experiments}
We now describe the datasets in (sec~\ref{sec:Datasets}), specify the implementation details in (sec~\ref{sec:Implementation}), and present our quantitative results in (sec~\ref{sec:Quant}).
\subsection{Datasets}\label{sec:Datasets}
Our problem is very particular in that we focus on multi-task learning for body affect. In the literature \cite{Kleinsmith_TAC2013,Zhang_arxiv2016} most of the datasets were either single task for activity recognition, not publicly available, too few instances, or only RGB-D without skeleton. We found two available datasets to evaluate our approach that are multi-task. The first dataset is the Body Affect dataset \cite{Ma_BRM2006}, collected using a motion capture sensor, which consists of a set of actors performing several actions with different affects. The second dataset is the Tower Game \cite{Salter_ACII2015}, collected using a Kinect sensor, which consists of an interaction between two humans performing a cooperative task, with the goal of classifying different components of entrainment. In the following subsections we describe the datasets.\\

\noindent{\bf Body Affect Dataset:} This dataset \cite{Ma_BRM2006} consists of a library of human movements captured using a motion capture sensor, annotated with actor, action, affect, and gender. The dataset was collected for studying human behavior and personality properties from human movement. The data consists of 30 actors (15 female and 15 male) each performing four actions (walking, knocking, lifting, and throwing) with each of four affect styles (angry, happy, neutral, and sad). For each actor, there are  40 data instances: 8 instances of walking (2 directions x 4 affects), 8 instances of knocking (2 repetitions x 4 affects), 8 instances of lifting (2 repetitions x 4 affects), 8 instances of throwing (2 repetitions x 4 affects), and 8 instances of the sequences (2 repetitions x 4 affects). For knocking and lifting and throwing there were 5 repetitions per data instances. Thus, the 24 records of knocking, lifting, and throwing contain 120 separate instances, yielding a total of 136 instances per actor and a total of 4,080 instances. We split dataset into 50\% training using 15 actors and 50\% testing using the other 15 actors.\\

\noindent{\bf Tower Game Dataset:} This dataset \cite{Salter_ACII2015} is a simple game of tower building often used in social psychology to elicit different kinds of interactive behaviors from the participants. It is typically played between two people working with a small fixed number of simple toy blocks that can be stacked to form various kinds of towers. The data consists of 112 videos which were divided into 1213 10-second segments indicating the presence or absence of these behaviors in each segment. Entrainment is the alignment in the behavior of two individuals and it involves simultaneous movement, tempo similarity, and coordination. Each measure was rated low, medium, or high for the entire 10 seconds segment. 50\% of that data was used for training and 50\% were used for testing. In this dataset we call each person's skeletal data a modality, where our goal is to model mocap-mocap representations.
\subsection{Implementation Details}\label{sec:Implementation}
For pre-processing the Tower Game dataset, we followed the same approach as \cite{Neverova_PAMI2015} by forming a body centric transformation of the skeletons generated by the Kinect sensors. We use the 11 joints from the upper body of the two players since the tower game almost entirely involves only upper body actions and gestures are done using the upper body. We used the raw joint locations normalized with respect to a selected origin point. We use the same descriptor provided by \cite{Neverova_ECCV2014,Zanfir_ICCV2013}. The descriptor consists of 84 dimensions based on the normalized joints location, inclination angles formed by all triples of anatomically connected joints, azimuth angles between projections of the second bone and the vector on the plane perpendicular to the orientation of the first bone, bending angles between a basis vector, perpendicular to the torso, and joint positions. As for the Body Affect dataset we decided to use the full body centric representation \cite{Zheng2015_CGF2015} for motion capture sensors resulting in 42 dimensions per frame.

For the Body Affect dataset we trained a three-task model for the following three tasks: Action (AC) $\in\{$Walking, Knocking, Lifting, Throwing$\}$, Affect (AF) $\in\{$Neutral, Happy, Sad, Angry$\}$, Gender (G) $\in\{$Male, Female$\}$.  The data is split into a training set consisting of 50\% of the instances, and a test set consisting of the remaining 50\%. For the Tower Game dataset we trained a three-task model for the following tasks,: Tempo Similarity (TS), Coordination (C), and Simultaneous Movement (SM), each in $\{$Low, Medium, High$\}$. The data is split into a training set consisting of 50\% of the instances, and a test set consisting of the remaining 50\%.

We tuned our model parameters. For selecting the model parameters we used a grid search. We varied the number of hidden nodes per layer in the range of $\{10,20,30,50,70,100,200\}$, as well as the auto-regressive nodes in the range of $\{5,10\}$, resulting a total of $2744$ trained models. The best performing model on the Body Affect dataset has the following configuration $v=42, h=30, v_{<t}=42 \times 10$ and the best performing model on the Tower Game dataset has the following configuration $v^{m}=84, h^{m}=30, v_{<t}^{m}=10 \times 84$ for each of the modalities and for the fusion layer in the Tower Game dataset $h^{1:M}=60, h=60, h_{<t}^{1:M}=10 \times 60$. 

Note that in our MT-CRBM model, the tasks are assumed conditionally independent given the hidden representation. Thus the number of parameters needed for the hidden-label edges is $H \cdot \sum_{k=1}^L Y_k$, where $H$ is the dimensionality of the hidden layer and $Y_k$ is the number of classes for task $k$. Contrast this to the number of parameters needed if instead the tasks are flattened as a Cartesian product, $H \cdot \prod_{k=1}^L Y_k$. Our factored representation of the multiple tasks uses only linearly many parameters instead of the  exponentially many parameters needed for the flattened representation.
\subsection{Quantitative Results}\label{sec:Quant}
\begin{figure*}[t]
	\centering
	\begin{minipage}{0.49\textwidth}
		\centering
		\includegraphics[width=0.98\textwidth]{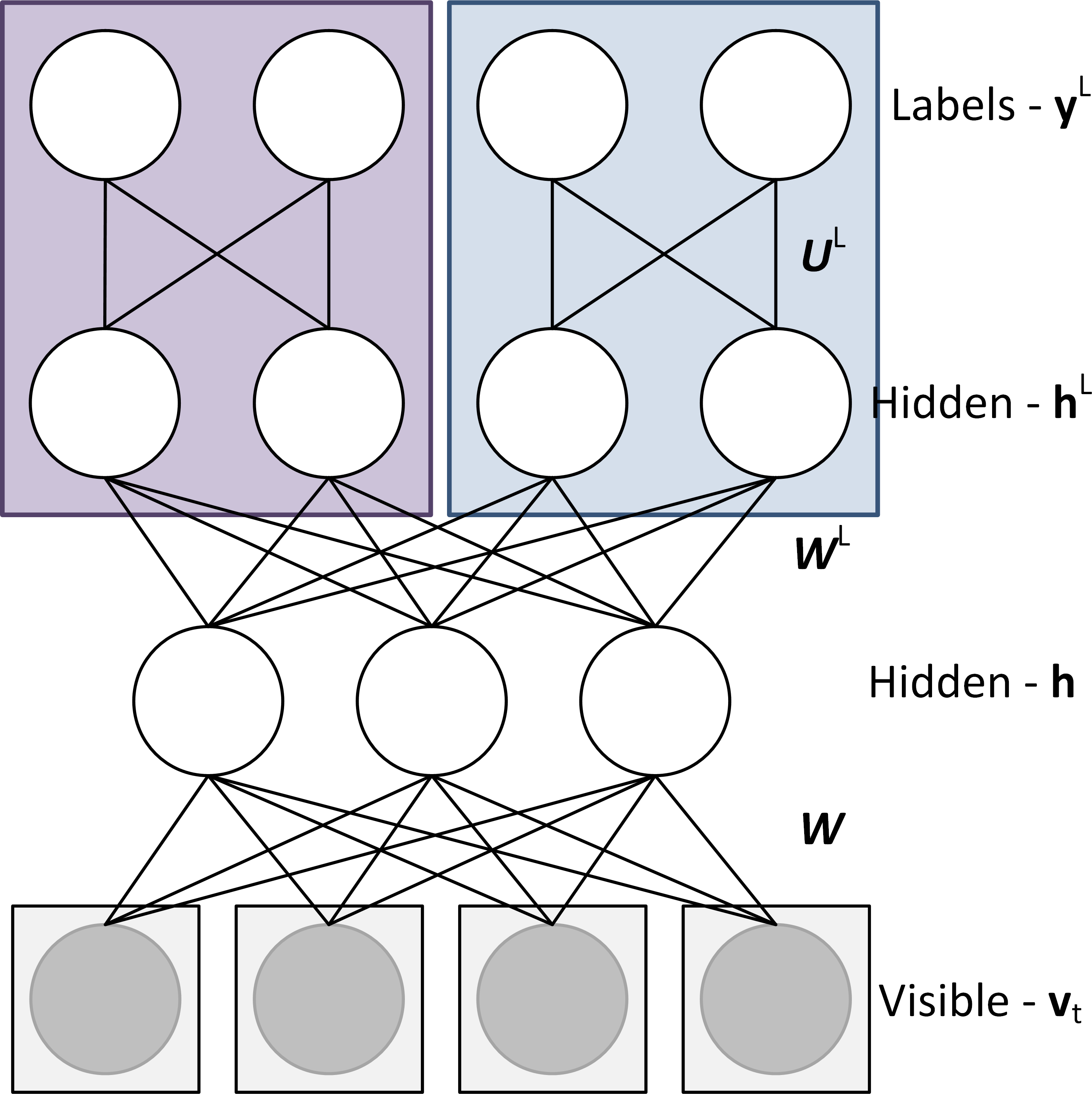}\\
		\scriptsize{(a) MT-CRBMs-Deep}	
		\vspace{48pt}	
	\end{minipage}	
	\hfill%
	\begin{minipage}{0.49\textwidth}
		\centering		
		\includegraphics[width=0.98\textwidth]{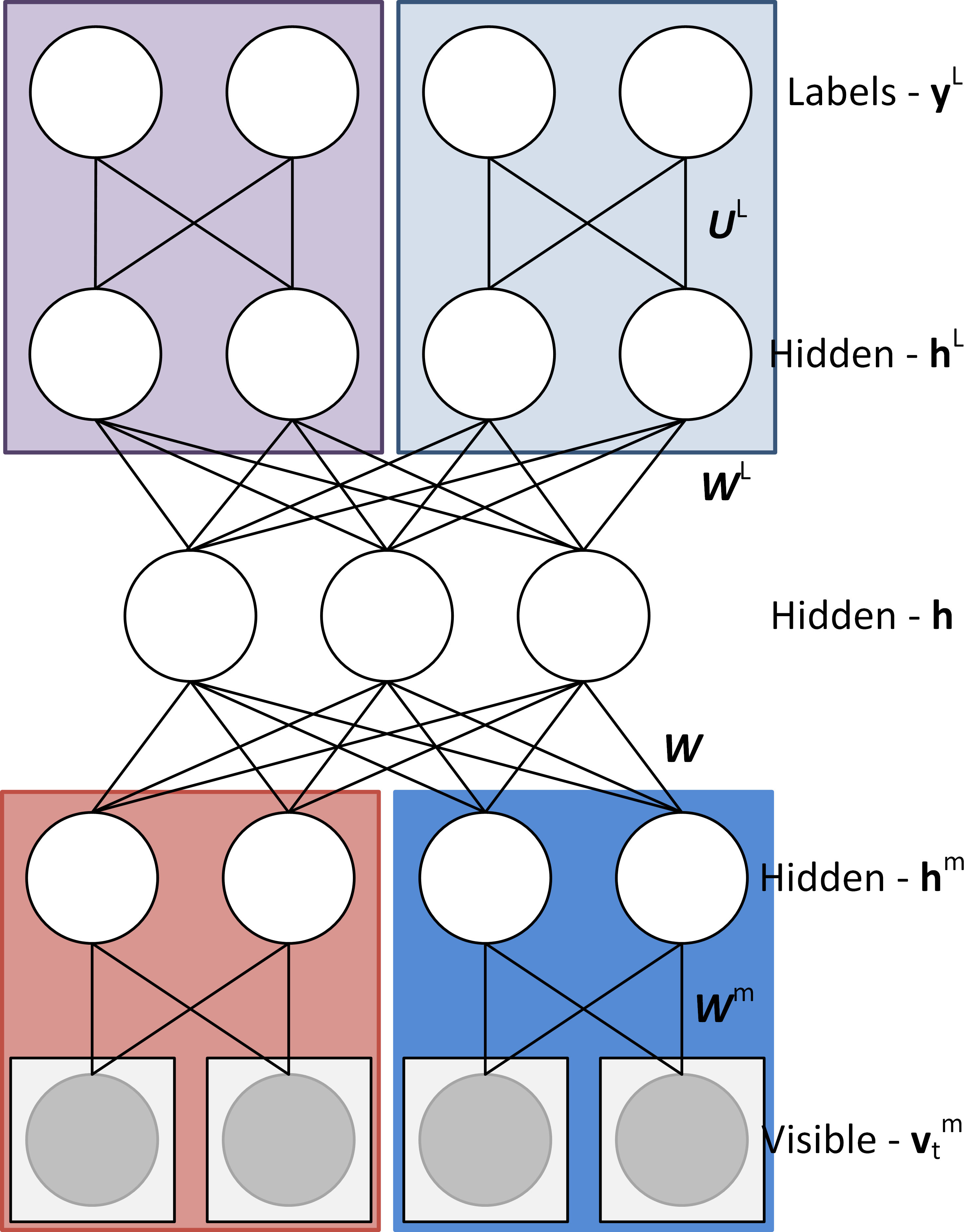}\\
		\scriptsize{(b) MTM-CRBM-Deep}		
	\end{minipage}	
	\caption{Deep variants of the models presented in sections~\ref{sec:MTCRBM} and ~\ref{sec:MTMCRBM}. (a) MT-CRBMs-Deep (b) MTM-CRBMs-Deep. The Deep variants add an extra representation layer for each of the tasks, which learns a task specific representation.}
	\label{fig:MTDeep}
\end{figure*}
We first define baselines and variants of the model, followed by the average classification accuracy results on the two datasets, and finally we provide some generative results, which we call {\it Morphing}, on the Body Affect dataset.\\

\noindent{\bf Baselines and Variants:} Since we compare our approach against the results presented in \cite{Salter_ACII2015} we decided to use the same baselines they used. They used SVM classifiers on a combination of features. {\it SVM+RAW:} The first set of features consisted of first order static and dynamic handcrafted skeleton features. The static features are computed per frame. The features consist of relationships between all pairs of joints of a single actor, and the relationships between all pairs of joints of both the actors. The dynamic features are extracted per window (a set of 300 frames). In each window, they compute first and second order dynamics of each joint, as well as relative velocities and accelerations of pairs of joints per actor, and across actors. The dimensionality of their static and dynamic features is (257400 D). {\it SVM+BoW100} and {\it SVM+BoW300:} To reduce their dimensionality they used, Bag-of-Words (BoW) (100 and 300 D) \cite{NieblesWF08,Zanfir_ICCV2013}. We also evaluate our approach using {\it HCRF} \cite{Wang_CVPR2006}. We define our own model's variants, {\it D-CRBMs} which is our single-task model presented in Section \ref{sec:DCRBM}, {\it MT-CRBMs} which is our multi-task model presented in Section \ref{sec:MTCRBM}, {\it MTM-CRBMs} the multi-modal multi-task model presented in Section \ref{sec:MTMCRBM} and {\it DM-CRBMs} an extension to the D-CRBMs to be multimodal similar to MTM-CRBMs. We also add two new variants\footnote{This model is initially prototyped by \cite{Goodfellow-et-al-2016-Book} in the deep learning book.} {\it MT-CRBMs-Deep} and {\it MTM-CRBMs-Deep} shown in Fig.\ref{fig:MTDeep}, which are a deeper version of the original models, by adding a task specific representation layer.\\

\noindent{\bf Classification:} For the Body Affect dataset, Table \ref{tab:BodyAffectResults3Task} shows the results of the baselines as well as our model and its variants. For the Tower Game dataset, Table \ref{tab:TowerGameResults} shows our average classification accuracy using different features and baselines combinations as well as the results from our models. We can see that the {\it MT-CRBMs-Deep} model outperforms all the other models for both cases, thereby demonstrating its effectiveness on predicting multi-task labels correctly. Furthermore, the {\it MTM-CRBMs-Deep} model outperforms all the SVM variants which used high dimensional handcrafted features, demonstrating its ability to learn a rich representation starting from the raw skeleton features. Note that only the {\it MTM-CRBMs} and {\it MTM-CRBMs-Deep} performed well on predicting the different tasks simultaneously with a relatively large margin better than the other models, using a shared representation that uses less parameters than our D-CRBMs model that treats all the labels flat.\\

\noindent{\bf Morphing:} Besides classifying action and affect, the {\it MT-CRBMs} model trained on the body affect dataset is also capable of generation. We demonstrate this by morphing a motion capture sequence of one affect to the same sequence with a different affect.  For example, we could morph a Neutral Walk into a Happy Walk. We morph a sequence by sweeping through its frames, updating the ${\bf v}_t$ vector of each frame in order. To update ${\bf v}_t$, we first compute the expected value of ${\bf h}_t$ given ${\bf v}_{<t}$, ${\bf v}_t$ and ${\bf y}_t^L$, then compute the expected value of ${\bf v}_t$ given ${\bf v}_{<t}$ and ${\bf h}_t$. The ${\bf v}_{<t}$ used is a linear blend of the original sequence and the newly generated sequence, so that the generated sequence retains the general shape of the original sequence. To evaluate the morphing process, we take a Neutral sequence for each action and each actor; morph it to a Happy, Sad, or Angry sequence of the same action type; and then compare the classifier probability of the target affect for the original Neutral sequence and the generated Happy, Sad, or Angry sequence. The average classifier probabilities before and after the morphing process for each action and target affect are shown in Table \ref{tab:BodyAffectMorphingResults}. Most of the classification probabilities for target affects increased as a result of morphing the Neutral sequences toward them, which means that our model was able to morph the sequences successfully.
\begin{table*}[t]
	\begin{center}
		\caption {Average Classification Accuracy on The Body Affect Dataset.} 
		\vspace{-3mm}
		\begin{tabular}{| l | c | c | c |}
			\hline
			Classifier (labels)	& AC(4)		& AF(4)		& G(2)				\\ \hline \hline
			Random Guess 		& 25.0 		& 25.0 		& 50.0				\\ \hline
			SVM+Raw     		& 35.6 		& 32.2 		& 65.1				\\ \hline
			SVM+BoW100 	 		& 41.3 		& 34.1 		& 71.4				\\ \hline
			SVM+BoW300\cite{Zanfir_ICCV2013}& 39.9 		& 32.8 		& 69.5 	\\ \hline
			HCRF\cite{Wang_CVPR2006}& 44.8 		& 34.7 	& 74.1 				\\ \hline
			D-CRBMs 		 		& 52.6 		& 30.7 		& 78.4			\\ \hline
			MT-CRBMs		 		& 53.5 		& 31.2 		& 78.2 			\\ \hline
			MT-CRBMs-Deep 		& 54.5 		& 32.7 		& 78.4 				\\ \hline
		\end{tabular}
		\label{tab:BodyAffectResults3Task} 
	\end{center}
	\begin{center}
		\caption{Average Classification Accuracy on The Tower Game Dataset.} 
		\begin{tabular}{| l | c | c | c | c |}
			\hline
			Classifier	(labels)	& TS (3)		& C (3)			& SM (3) 	\\ \hline \hline
			Random Guess 			& 33.3			& 33.3 			& 33.3 	 	\\ \hline
			SVM+Raw  \cite{Salter_ACII2015}    			& 59.3			& 52.2 			& 39.5	\\ \hline
			SVM+BoW100 \cite{Salter_ACII2015}	 			& 65.6			& 55.8 			& 44.3 \\ \hline
			SVM+BoW300 \cite{Zanfir_ICCV2013}& 54.4	& 47.5 			& 42.8 		\\ \hline
			HCRF\cite{Wang_CVPR2006}& 67.2			& 58.8 			& 44.5 		\\ \hline
			DM-CRBMs  	 			& 76.5			& 62.0		 	& 49.2 		\\ \hline
			MTM-CRBMs   	 		& 86.2			& 70.0		 	& 63.5 		\\ \hline
			MTM-CRBMs-Deep			& 87.2			& 70.0		 	& 72.8 		\\ \hline
		\end{tabular}
		\label{tab:TowerGameResults} 
	\end{center}
	\begin{center}
		\caption {Affect Classification Before and After Morphing} 
		\begin{tabular}{| l || c | c || c | c || c | c |}
			\hline
			AC/AF		  & Happy-Before & Happy-After & Sad-Before & Sad-After & Angry-Before & Angry-After \\ \hline \hline
			Knock         & 32.0     	 & 33.3    	   & 14.6  		& 15.2 		& 15.2    	   & 15.5   	 \\ \hline
			Lift          & 18.0     	 & 19.7    	   & 11.3  		& 21.1 		& 9.1    	   & 14.7   	 \\ \hline
			Throw         & 12.9     	 & 13.9    	   & 8.1  		& 13.3 		& 6.9    	   & 8.6   	 \\ \hline
			Walk          & 36.7     	 & 46.9    	   & 4.3  		& 4.3 		& 4.1    	   & 8.0   	 \\ \hline
		\end{tabular}
		\label{tab:BodyAffectMorphingResults} 
	\end{center}
\end{table*}
\section{Conclusion and Future Work}\label{sec:Conclusion}
We have proposed a collection of hybrid models, both discriminative and generative, that model the relationships in and distributions of temporal, multimodal, multi-task data. An extensive experimental evaluation of these models on two different datasets demonstrates the superiority of our approach over the state-of-the-art for multi-task classification of temporal data. This improvement in classification performance is accompanied by new generative capabilities and an efficient use of model parameters via factorization across tasks.

The generative capabilities of our approach enable new and interesting applications, such as the demonstrated sequence morphing. A future direction of work is to further explore and improve these generative applications of the models.

The factorization of tasks used in our approach means the number of parameters grows only linearly with the number of tasks and classes. This is seen to be significant when contrasted with a single-task model that uses a flattened Cartesian product of tasks, where the number of parameters grows exponentially with the number of tasks. Our factorized approach makes adding additional tasks a trivial matter.
\section*{Acknowledgments}
This research was partially developed with funding from the Defense Advanced Research Projects Agency (DARPA) and the Air Force Research Laborotory (AFRL). The views, opinions and/or findings expressed are those of the authors and should not be interpreted as representing the official views or policies of the Department of Defense or the U.S. Government
\clearpage
\bibliographystyle{splncs}
\bibliography{References}
\end{document}